%% file: iclr2026_conference.tex
\documentclass{article} 
\usepackage{iclr2026_conference,times}

\input{math_commands.tex}

\usepackage{hyperref}
\usepackage{url}
\usepackage{comment}
\usepackage{graphicx}
\usepackage{pifont}
\usepackage{booktabs}
\usepackage{listings}
\graphicspath{{./Images/}}
\usepackage{colortbl}
\usepackage{wrapfig}
\usepackage{multirow}
\usepackage{tabularx}
\usepackage{adjustbox}
\usepackage{subcaption}
\usepackage{amssymb}
\usepackage{diagbox}
\usepackage{amsfonts}       
\usepackage{nicefrac}       
\usepackage{microtype}      
\usepackage{varwidth}
\usepackage{xcolor}  
\title{\textsc{Shanks}: Simultaneous Hearing and Thinking for Spoken Language Models}

\author{%
  Cheng-Han Chiang\textsuperscript{1,2}\thanks{Work done during an internship at Microsoft GenAI. \texttt{dcml0714@gmail.com}} \quad
  Xiaofei Wang\textsuperscript{2}\thanks{Correspondence: \texttt{xiaofei.wang@microsoft.com}} \quad
  Linjie Li\textsuperscript{2} \quad
  Chung-Ching Lin\textsuperscript{2} \quad
  Kevin Lin\textsuperscript{2} \quad\\
  \textbf{Shujie Liu}\textsuperscript{2} \quad
  \textbf{Zhendong Wang}\textsuperscript{2} \quad
  \textbf{Zhengyuan Yang}\textsuperscript{2} \quad
  \textbf{Hung-yi Lee}\textsuperscript{1} \quad
  \textbf{Lijuan Wang}\textsuperscript{2}\\[1ex]
  \textsuperscript{1}National Taiwan University \quad
  \textsuperscript{2}Microsoft
}

\iclrfinalcopy 
\begin{document}

\maketitle

\begin{abstract}
Current large language models (LLMs) and spoken language models (SLMs) begin thinking and taking actions only \textit{after} the user has finished their turn.
This disables the model from interacting with the user during the user's turn and can lead to a high response latency for waiting for the model to think.
Consequently, thinking \textit{after} receiving the full input is not suitable for speech-to-speech interaction, where real-time and low-latency interaction is important.
We address the above issue by drawing inspiration from the fact that humans can naturally \textit{``think while listening''}. 
In this paper, we propose \textbf{\textsc{Shanks}}, a general inference framework that enables SLMs to generate unspoken chain-of-thought reasoning when listening to the user input.
\textsc{Shanks} streams the input speech in fixed-duration chunks and, as soon as a chunk is received, generates unspoken reasoning based on all previous speech and reasoning, while the user continues speaking.
\textsc{Shanks} uses unspoken reasoning to determine whether to interrupt the user and make tool calls to complete the task.
We demonstrate that \textsc{Shanks} enhances the real-time user-SLM interaction in two scenarios:
(1) When the user is presenting their step-by-step solution to a math problem, \textsc{Shanks} can listen to and reason over the user's speech and make an interruption when the user makes a mistake.
\textsc{Shanks} interrupts the user 37.1\% more accurately compared with a baseline that interrupts the user without thinking.
(2) In a tool-augmented dialogue scenario, where the model needs to make tool calls to achieve the user's request, \textsc{Shanks} can complete 56.9\% of the tool calls before the user even ends their turn.
Overall, \textsc{Shanks} is a step toward models that keep thinking throughout the conversation, not only after a turn ends.
Animated illustrations of \textsc{Shanks} can be found at \url{https://d223302.github.io/SHANKS/}. 
\end{abstract}

\begin{figure}[ht!]
\includegraphics[clip, trim = 20px 16px 20px 16px,width=0.98\linewidth]{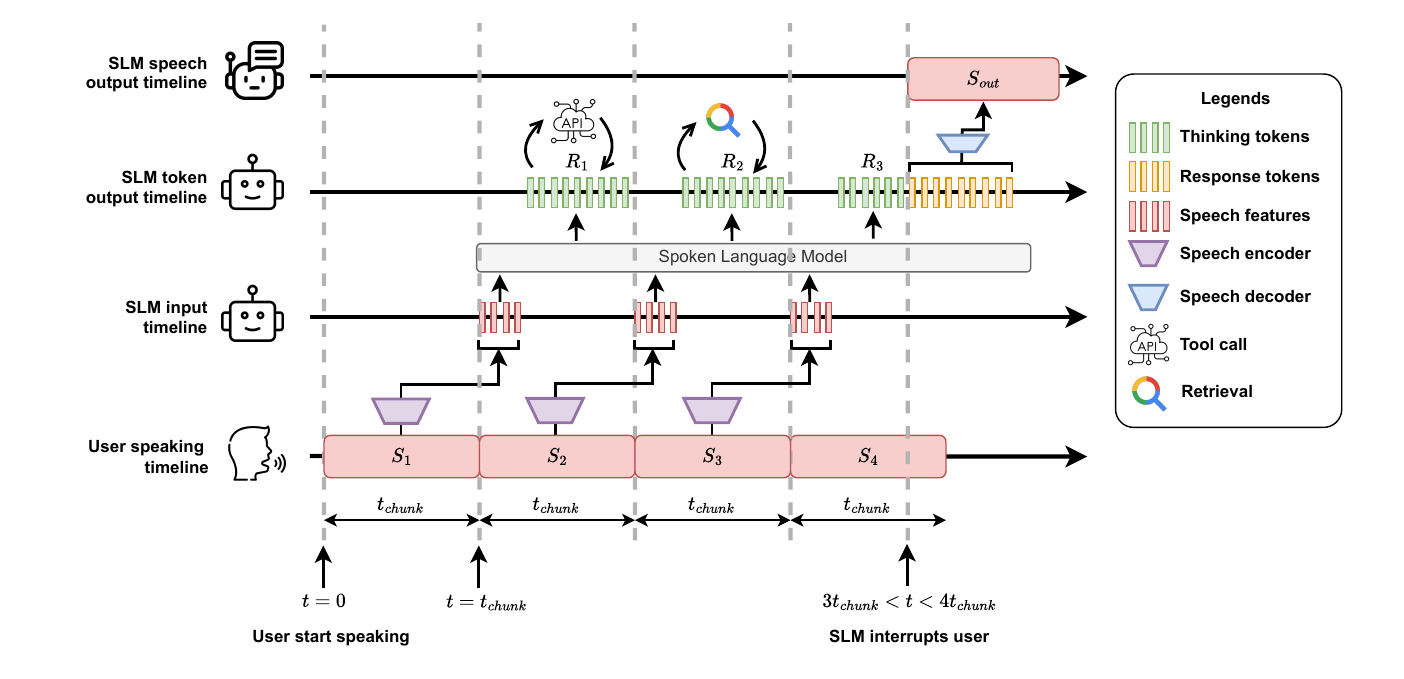}
\caption{
The timing diagram of \textsc{Shanks}.
As the user speaks, their speech is segmented into chunks for every $t_{chunk}$ seconds and streamed to the SLM.
After receiving an input chunk, \textsc{Shanks} generates the thinking tokens, which might include calling external tools or determining to interrupt the user.
When the user is speaking the $i$-th speech chunk $S_i$, \textsc{Shanks} generates the $(i-1)$-th thinking chunk $R_{i-1}$, achieving thinking while listening.
When the current speech chunk $S_{i}$ is fully spoken by the user, \textsc{Shanks} stops the thinking for $R_{i-1}$, adds the latest speech $S_{i}$ and the previous reasoning $R_{i-1}$ to its context, and begins the $i$-th thinking chunk $R_{i}$.} 
    \label{fig:fig1}
\end{figure}

\section{Introduction}
\label{Section: Introduction}

In recent years, the \textit{thinking} process has been used to improve Large Language Models (LLMs), where the LLM first generates a \textit{hidden} chain-of-thought (CoT) reasoning~\citep{cot,kojima2022large} invisible to the users, and then generates the final output response~\citep{o1,guo2025deepseek}.
This thinking process improves LLMs on reasoning-intensive tasks, including mathematics~\citep{math500}, coding~\citep{humaneval}, and questions that involve significant domain knowledge~\citep{rein2024gpqa}.
However, current reasoning LLMs only start to think \textit{after} receiving the complete user input, which is reasonable for turn-based interactions, i.e., the model processes the user's message after it is fully composed and sent.

In contrast, human behavior in \textit{spoken} communication is different. 
Humans naturally think \textit{while} listening, far before the speaker finishes their turn~\citep{bogels2015neural,corps2018early,bogels2018planning}.
Thinking during listening offers two key advantages: 
(1) It enables timely and well-founded reactions, including interruption, even before the speaker's turn ends.
(2) It reduces response latency by allowing answer preparation to begin before the speaker finishes speaking. 
Motivated by these observations, we propose a method to enable spoken language models (SLMs) to think while listening to input speech.

In this paper, we introduce \textsc{Shanks}: \textbf{\underline{S}}imultaneous \textbf{\underline{H}}earing \textbf{\underline{a}}nd Thi\textbf{\underline{nk}}ing with Chunked Input \textbf{\underline{S}}peech.
\textsc{Shanks} is a general inference framework for both end-to-end (E2E) and cascade SLMs to achieve thinking while listening.
At inference time, \textsc{Shanks} processes the user input in a fixed-size chunk.
Once a chunk of speech input is received, \textsc{Shanks} generates a chunk of unspoken thinking tokens based on all previous input speech chunks and previous thinking chunks.
\textsc{Shanks} alternates between receiving the input speech chunk and generating an unspoken CoT reasoning when the user is still emitting the next speech chunk, achieving the \textit{\textbf{thinking while listening}}.
During the thinking process, \textsc{Shanks} can decide to interrupt the user or make tool calls to prepare for the final spoken response.
The inference workflow of \textsc{Shanks} is depicted in Figure~\ref{fig:fig1}.

We use two scenarios to show how \textsc{Shanks} can improve real-time user-SLM interaction.
First, we study a scenario where the user first describes a math question and then describes their step-by-step solution.
\textsc{Shanks} can listen to the user's problem-solving process and perform internal thinking in the meantime to interrupt the user when the user makes a mistake in their solution.
This scenario has great potential in educational use cases, where the SLM serves as a tutor to guide the student.
Compared to a baseline that makes an interruption without thinking, \textsc{Shanks} interrupts 71\% more when the user makes a mistake, while the interruption made by \textsc{Shanks} is 37.1\% more valid.

Next, we focus on a task-oriented dialogue setting, where the user requests the SLMs to assist with a travel plan, and the model needs to call Booking.com APIs to complete the user's request and respond to the user.
Without \textsc{Shanks}, all the tool calls can only be made after the user’s speech ends, resulting in a higher response latency.
We use \textsc{Shanks} to make tool calls when the user is still speaking.
\textsc{Shanks} enables the model to successfully complete 56.9\% of API calls while the user is still speaking, reducing the latency of the final response.

We summarize our contribution as follows:
\begin{enumerate}
    \item We propose \textsc{Shanks}, a general framework for SLMs that enables \textit{thinking while listening}.
    To the best of our knowledge, we are the first to explore generating unspoken CoT reasoning when the user is still speaking.
    \item We show that \textsc{Shanks} can interrupt the user more accurately compared to a baseline that interrupts without thinking.
    \item Using \textsc{Shanks}, the SLM can successfully make tool calls \textit{before the user even finishes talking}.
\end{enumerate}

\section{Method: Simultaneous Hearing and Thinking with Chunked Input Speech}
\label{Section: Method: Thinking While Listening}

Current LLMs and SLMs only start to think after the user's input is completed.
In contrast, humans can think while listening~\citep{bogels2015neural,corps2018early,bogels2018planning}, where we reason over what we just heard, guess what the speaker might be up to, and prepare the necessary ingredients to cook up a good response. 
Thinking while listening allows us to interact with the speaker better when the speaker is still speaking.
In this section, we introduce \textsc{Shanks}, a general framework to make SLMs capable of thinking while listening.
Here, we only discuss the basic form of \textsc{Shanks}, and we defer the more advanced usages, including interruption or tool call, to later sections.

\subsection{Inference}
\label{Subsubsection: Inference}

During inference, \textsc{Shanks} requires that the user's input speech comes in a streaming manner.
\textsc{Shanks} processes the streaming user input speech by a fixed chunk size of $t_{\rm chunk}$ seconds.
We use $S_i$ to denote the $i$-th user input speech chunk, where $S_i$ is an audio chunk of $t_{\rm chunk}$ seconds, except for the last chunk $S_{N}$, which may be shorter.
When the user is still speaking, \textsc{Shanks} alternately takes the user speech $S_{i}$ and generates the thinking chunks $R_{i}$ conditioning on all previous user speech and all previous thinking chunks.
The thinking chunks $R_i$ are \textbf{not spoken by the SLM}; they only serve as the internal reasoning process of the SLM.

Here, we walk through what happens for \textsc{Shanks} during inference.
The following contents are best read with Figure~\ref{fig:fig1}.
At $t=0$, the user begins to talk.
When $t=t_{\rm chunk}$, the user speech from $0$ to $t_{\rm chunk}$, i.e., $S_1$, is sent to the SLM.
Here, we append a special token \texttt{[EOPA]} (end of partial audio) after $S_1$ to let the SLM know that this is the end of a chunk of \textit{partial} user speech.
Based on $S_1$, the SLM generates the first thinking chunk $R_1$.
A thinking chunk is enclosed in two special tokens \texttt{<think>} and \texttt{</think>}.
The SLM generates $R_1$ during the interval $t=t_{\rm chunk}$ to $t=2t_{\rm chunk}$, and the user is still speaking the second chunk $S_2$ at the same time.

Since $t_{\rm chunk}$ is the time for the SLM to generate its thinking, the duration of $t_{\rm chunk}$ cannot be selected too small; otherwise, the SLM may not be able to produce meaningful thinking chunks.
The number of thinking tokens in $R_i$ is restricted to no more than $t_{\rm chunk}\times n_{\rm tps}$, where $n_{\rm tps}$ is the number of tokens the model can generate per second.
Unless specified, we select $t_{\rm chunk}=4.0s$ in our paper; a 7B model can generate around 320 tokens on a single A100 GPU in this duration.
At the end of $t_{\rm chunk}$, if the thinking chunk has not finished generating, i.e., the \texttt{</think>} token has not been emitted, we directly stop generating and append the \texttt{</think>} token at the end of the thinking chunk.

At $t=2t_{\rm chunk}$, we take the freshly obtained user speech chunk $S_2$ (from $t=t_{\rm chunk}$ to $t=2t_{\rm chunk}$) and pass this chunk to the SLM, and again appending the \texttt{[EOPA]} after this chunk to generate the next thinking chunk. 
(Assume that the user still has not ended their turn at $t=2t_{\rm chunk}$.)
When generating the next thinking chunk  $R_2$, the SLM conditions on $S_1$, $R_1$, and $S_2$.
The SLM will continue the process of taking user input speech chunks and generating the thinking chunks until the user ends their speech in the $N$-th chunk of speech, $S_N$.
After the user ends their speech, we feed the last speech chunk $S_{N}$ into the SLM, while this time we append a different special token \texttt{[EOA]} (end of audio), indicating that the user's speech has ended.
Based on $S_N$ and all the previous interleaved speech/thinking chunks $\{S_1, R_1,\cdots,S_{N-1},R_{N-1}\}$, the SLM generates the thinking chunk $R_N$ and then generates a final response chunk $O$.
Only the final response $O$ will be spoken by the SLM.

Since \textsc{Shanks} chunks the user input using a fixed-duration chunk $t_{\rm chunk}$, the model's thinking will lag behind the user's speech by at least $t_{\rm chunk}$ seconds.
If the user's speech is less than $t_{\rm chunk}$, \textsc{Shanks} cannot think while listening.
However, since long speech can easily happen in real-world interaction, this limitation might not be a significant weakness of \textsc{Shanks}.

\subsection{Training}
\label{Subsubsection: Training}

\begin{figure*}[t!]

\centering
\includegraphics[clip, trim = 27px 3px 15px 3px,width=0.98\linewidth]{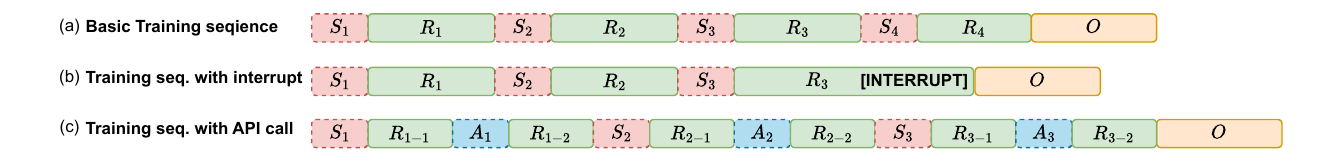}
\caption{Illustration of the training data. 
$S_i$: the speech for the $i$-th user speech chunk; $R_i$: the $i$-th thinking block after $S_i$; $O$: the final response block; $A_i$: the API call responses after the speech chunk $S_i$.
Blocks in dashed lines do not contribute to the training loss, while blocks in solid lines are included for loss calculation.
(a) The general training sequence: Alternating between user speech block and SLM thinking token chunks (Section~\ref{Subsubsection: Training}), followed by a final response chunk.
(b) Training data with interruption: Alternating between user speech blocks and the thinking token chunks, while the last thinking chunk includes a special token \texttt{[INTERRUPT]}.
(c) Training data with API calls: Similar to (a), while each thinking chunk may be separated into two blocks $R_{i-1}$ and $R_{i-2}$ by the API call response $A_i$.}
\label{fig:training_data.pdf}
\end{figure*}

During inference, \textsc{Shanks} requires the SLM to generate thinking chunks based on all previous user input speech chunks and the model's own thinking chunks.
During training, we prepare datasets to make the SLM learn this behavior.
Assume that we have a complete user speech $S$, we can segment it into $N$ chunks $\{S_1,\cdots,S_N\}$ with a fixed duration $t_{\rm chunk}$.
Next, assume that we use some method to obtain the thinking chunks $\{R_1,\cdots,R_N\}$ and the output response $O$; we will explain how to obtain them in later sections.
After preparing the training data, we use the standard language modeling cross-entropy loss to train the SLM to predict $R_1$ given $\{S_1\}$, predict $R_2$ given $\{S_1,R_1, S_2\}$, etc., and predict $R_N$ and $O$ given $\{S_1,R_1,\cdots,S_{N-1}\}$.
A full training sequence is depicted in Figure~\ref{fig:training_data.pdf}(a).

\section{Task Introduction}
\label{Section: Task Introduction}

After introducing the basics of \textsc{Shanks}, we use two tasks to demonstrate how \textsc{Shanks} can be applied.
In this section, we explain the setup of the two scenarios and how \textsc{Shanks} works at inference time.

\subsection{Scenario 1: Interrupting User Turn}
\label{Section: Application 1: Thinking While Listening for Interrupting User Turn}

In the first scenario, we aim to use \textsc{Shanks} to make SLMs able to interrupt the user when the user is saying something wrong.
The significance of this application lies in its potential in educational use cases, where the SLMs can serve as a tutor and listen to the speaker, a student, describing how they solve a problem.
The SLM can make a timely interruption to let the student know that they are making a mistake, allowing them to correct it as early as possible.

Interruption is related to the full-duplex ability of spoken language models~\citep{lin2025fullduplexbenchbenchmarkevaluatefullduplex}.
While most prior works on full-duplex SLMs focus on user interrupting SLMs, we focus on the reverse scenario: SLM interrupting users.
As an important note, we do not advocate that it is good to have a model that interrupts the user.
Some users might find it annoying and unpleasant when interrupted by an SLM.
Our goal is a modeling mechanism that enables interruption, leaving model deployer policies and users to decide when (or whether) to turn it on.

\subsubsection{Task Description}
\label{Subsection: App 1 Task Description}

We explain the precise task we are studying.
In this task, the user describes a math problem and then solves the problem.
The user's solution does not simply state the answer; the user describes a step-by-step problem-solving process, which might be correct or wrong.
The SLM needs to interrupt the user when the user is making a mistake, and not to interrupt the user when the solution is correct.

As this is a novel task and there is no available data, we built the evaluation data ourselves.
First, we construct the user speech.
A single user speech includes (1) a math question and (2) a step-by-step solution.
We source the math questions from the testing data of GSM8K~\citep{cobbe2021training}, a grade-school math word problem dataset commonly used for evaluating mathematical reasoning ability~\citep{cot,kojima2022large,wang-etal-2023-plan}.
Next, we use two LLMs, Llama-2-7B~\citep{touvron2023llama} and Llama3.1-8B~\citep{grattafiori2024llama}, to generate step-by-step answers for those questions, and use GPT-4o~\citep{gpt4o} to determine if the answer generated by the two models matches the ground truth answer in the dataset.
We select these two models since they can generate CoT reasoning to solve the math problem, and their performance on GSM8K is very different: Llama-2-7B is a weaker model and prone to generating wrong solutions, while Llama-3.1-8B is a stronger model, which can generate more accurate solutions.

After we have the texts for the step-by-step solution, we convert them into speech.
We use GPT-4o to rewrite the answers generated by the two Llama models to make the solution more colloquial.
Next, we concatenate the original question, the colloquial step-by-step answer, and prepend a prefix \textit{"I want to solve the following question."} to form the transcription of a testing instance.
We use GPT-4o-mini-TTS~\citep{4ominitts} to synthesize the speech.

The final testing dataset includes 1280 instances with correct solutions and 1140 with incorrect solutions.
We call the former the "\textit{correct subset}" and the latter the \textit{"incorrect subset}. 
The average duration of the user's speech is around 49.25 seconds.

\subsubsection{Training Data for Interruption}
\label{Subsection: Training Data for Interruption}

To teach the model to think while listening and determine whether to interrupt, the training data in this task include two types of instances:
(1) The user provides a correct step-by-step solution to the question, and the model does not interrupt the user during the user's speech. 
After the user finishes the speech, the output response acknowledges the correctness of the answer.
(2) The user's turn unfolds an erroneous problem-solving process, and the model interrupts the user when the user makes the first mistake and clearly explains what is wrong.

To construct such a training dataset, we use the math questions in Tulu3-Persona-Math-Grade~\citep{lambert2024tulu} to construct the user speech $S$ (including the question and the step-by-step solution) following the previously described procedure, and then segment the speech by a fixed duration $t_{\rm chunk}=4$ seconds to obtain $\{S_{1},\cdots,S_{N}\}$.

Next, we use GPT-4o to generate the thinking chunk $R_i$.
When generating the $i$-th thinking chunk $R_i$, the input to GPT-4o includes the transcriptions of all previous user speech chunks $\{S_1,\cdots,S_{i}\}$ and all previous thinking chunks $\{R_1,\cdots,R_{i-1}\}$.
GPT-4o is required to do the following in the thinking chunk:
(1) Track the information already known and calculate intermediate variables when they are available.
(2) Identify if any errors are in the current user's transcription.
If there is an error, GPT-4o should generate a \texttt{[INTERRUPT]} token at the end of the thinking chunk, indicating that the user should be interrupted.
We give GPT-4o four in-context examples to allow GPT-4o to understand the task.
The prompt to GPT-4o is given in Table~\ref{tab:thinking-while-listening-1} and \ref{tab:thinking-while-listening-2} in the Appendix.

After generating the thinking chunks, we generate the final output response $O$.
For the user speeches with an error-free solution, the output response simply needs to let the user know that their solution is correct.
We prompt GPT-4o to generate the final response based on the transcription of the full user speech and all previous thinking chunks.
Now, we can form a training data sequence by interleaving $S_i$ and $R_i$ and then appending $O$ in the end.

For those user speeches with a wrong solution, the output response will be an interruption to the user's speech.
Assume that based on our previous process for generating the $R_i$'s, GPT-4o decides to interrupt the user after the user speech chunk $S_k$, i.e., the thinking chunk $R_k$ includes the interruption token \texttt{[INTERRUPT]}.
To generate a response for interruption, we give GPT-4o the user's speech up to the $k$-th user speech chunk and all the previous thinking, and ask GPT-4o to generate a response $O$ to interrupt the user.
The interruption should be precise on what error is made by the user and how to correct it.
After this process, we can interleave $S_1$ to $S_k$ with $R_1$ to $R_k$ and append $O$ in the end to form a training sequence.
A figurative illustration of this training instance is shown in Figure~\ref{fig:training_data.pdf}(b).
Note that in the last thinking chunk $R_{k}$, there will be a special token \texttt{[INTERRUPT]}, indicating that the model is going to interrupt the user.

\subsubsection{Inference and Evaluation}
\label{subsection: Evaluation interruption}

During inference, we stream the user speech to the SLM by a fixed chunk size $t_{\rm chunk}$, and follow the inference procedure elaborated in Section~\ref{Subsubsection: Inference}.
If the SLM generates the special token \texttt{[INTERRUPT]} in a thinking chunk $R_k$ and outputs a response chunk $O$ (when the user is emitting speech chunk $S_{k+1}$), we convert the response token into speech to interrupt the user.
Once an interruption happens, the future user speech chunks will not be streamed to the SLM.

We evaluate a model on the testing dataset constructed in Section~\ref{Subsection: App 1 Task Description}.
We use the following metrics:
\begin{enumerate}
    \item \textbf{Interrupt ratio}: The ratio of total interrupted instances among the total instances.
A good model should have a low interrupt ratio on the correct subset and a high interrupt ratio on the wrong subset.

\item \textbf{Valid interrupt ratio}: This is used to evaluate whether an interruption made by the model is valid, and the valid interrupt ratio is defined as the ratio of \textit{valid} interruptions among the total interruptions. 
To judge whether an interruption response $O$ is valid or not, we use LLM-as-a-judge~\citep{chiang-lee-2023-large,zheng2023judging}.
We give the judge LLM (GPT-4o) the transcription of the user input until the time of interruption\footnote{If the interruption happens in $R_i$, the user is currently speaking the $(i+1)$-th speech chunk $S_{i+1}$, as the thinking $R_i$ happens simultaneously when the user says $S_{i+1}$. Consequently, we also feed the transcription of $S_{i+1}$ into the judge model when determining whether the interruption is valid. Nevertheless, we do not find the results to differ significantly if we only give the judge model up to the speech chunk $S_i$.} and output response $O$ from the model, and ask the LLM judge to determine if the model's interruption response $O$ correctly interrupts the user when there are unclear or mistakes in the user's speech.
The prompt given to the judge model is shown in Table~\ref{tab:assistant-interruption-judgement} in the Appendix.

\item \textbf{Interruption latency}: The time of the model interruption compared to \textit{the time when the first error happens in the user's speech}, denoted as $t_{\rm error}$.
This metric is only used to evaluate the incorrect subset.
For samples in the incorrect subset, we use GPT-4o to determine $t_{\rm error}$.
The details on determining $t_{\rm error}$ are included in Appendix~\ref{subsection: Evaluation Details for Interruption}.
Assume that the model interrupts the user at $t_{\rm interrupt}$, then the interruption latency is calculated as $t_{\rm interrupt} - t_{\rm error}$, where $t_{\rm interrupt}$ is the time when the model starts to emit the first package of the speech for the interruption output $O$.

\end{enumerate}

\subsection{Scenario 2: Making Tool Calls When Listening}
\label{Section: Application 2: API Call When Listening to Reduce Response Latency}

In the previous scenario, we have introduced \textsc{Shanks} can be used to think when listening to interrupt the user when necessary.
However, the "\textit{thinking}" of a model can not only include the CoT reasoning generated by the model itself; the model can use external tools in their thinking process to complete the user's request and improve the answer quality~\citep{schick2023toolformer,qin2024toolllm,qin2024toollearningfoundationmodels,gao2024retrievalaugmentedgenerationlargelanguage,wang2024retrieval}.
In the second scenario, the SLMs will use \textbf{\textit{external tools}} in their thinking process to achieve the user's request.

This kind of \textbf{\textit{tool-augmented generation}} has been widely explored in text-in-text-out LLMs~\citep{qin2024toollearningfoundationmodels}.
Given a user request, a tool-augmented LLM selects relevant tools such as calculators~\citep{schick2023toolformer}, search engines~\citep{luo-etal-2023-search}, and other APIs~\citep{qin2024toolllm}, and integrates the tool execution results into its reasoning process.
However, current tool-augmented LLMs begin calling the tools after the full user input is received, which adds delay while the model invokes tools, waits for results, parses them, and composes a response.
This delay may be acceptable in text-only settings, but in spoken dialogue, this latency breaks conversational flow.

In the second task, we will use \textsc{Shanks} to make tool calls when the user is still speaking, thus reducing the response latency due to making tool calls.
As a proof of concept, we consider a task-oriented dialogue where the SLM serves as a travel-agency agent and is given a set of APIs it may call to complete the task.
For example, the user might say, \textit{"Help me check the details of the cheapest flight from Hangzhou to Seoul on December 10, 2024, and the car rental information near Seoul airport.}"
The SLM makes API calls to resolve airport names or codes, search for flights, and then query car-rental options, and finally composes the results into a single reply.

Given a user query like the example above, once the destination and date are clear, the flight search API can be called even when the user is only halfway through speaking.
This is where \textsc{Shanks} can be useful: processing partial user input and performing early actions.
Next, we formally introduce the task we study and how we evaluate it.

\subsubsection{Task Description}
\label{Subsection: API Call Task Description}
To show that SLMs can make tool calls when listening to the user request, we adopt ComplexFuncBench~\citep{zhong2025complexfuncbench}, an evaluation dataset that assesses LLMs' ability to make multi-step API calls.
An instance in ComplexFuncBench includes a user query in text specifying some requirements for a travel plan and a list of tool descriptions that are required to complete the task.
Some example tools include APIs for searching hotels or flights.

An evaluated model needs to make relevant API calls and provide the resulting information to the user.
A single user query typically requires multiple API calls, and these calls may have dependencies in which the output of one call becomes an argument to a subsequent call, so the call order matters and some calls may not be run in parallel.
For each instance, the dataset provides the ground truth API calls and their corresponding API responses, which can be used to evaluate whether the model's API call is correct.
To adapt ComplexFuncBench to our spoken-dialogue setting, we use GPT-4o-mini-TTS to synthesize the user's spoken query from the text instructions from ComplexFuncBench.

\subsubsection{Training Data for Tool Call}
\label{Subsection: Training Data for API Call}

To train \textsc{Shanks} to perform Tool calls when listening, we need to teach the model to make API calls in their thinking process $R_i$ based on user input speech chunk $S_i$.
We split half of the instances in ComplexFuncBench to construct the training data and the other half as the testing data.\footnote{ComplexFuncBench is originally designed as an evaluation dataset. Here, we train the model directly on this dataset since the models we use were not trained to perform tool call. Since our training data has the \textbf{exactly same} distribution as the testing data, our results should not be compared with other models that are not trained on this dataset.}
The speech chunks $S_i$ in the training data can simply be obtained from segmenting the audio of the user query speech.

The next step is to construct the thinking chunks $R_i$.
In this task, \textbf{the thinking chunk $R_i$ is the API calls and call responses}.
For each user query, ComplexFuncBench already provides the ground truth API calls to complete the task, and we only need to determine which API calls, among the ground truth API calls, can be made after a speech chunk $S_i$.
To determine which API calls can be made in $R_i$, recall that a thinking chunk $R_i$ is based on the user speech up to $i\times t_{\rm chunk}$ seconds, so as long as the speech up to $i\times t_{\rm chunk}$ provides the information to make an ground truth API call, that API call can be included in $R_i$.
We follow the above idea and prompt GPT-o1~\citep{o1} with the transcription of the user speech, the time alignment of each word in the user's speech, and the ground truth APIs, and ask GPT-o1 to determine the earliest time an API call can be made.
The prompt is shown in Table~\ref{tab:earliest-tool-call} in the Appendix.
Based on the above process, we can determine which ground truth APIs should be included in which thinking chunk $R_i$. 
A thinking chunk $R_i$ can have more than one API call and response. 
If no API calls can be made in $R_i$, we put a template message that says there are no additional tool calls that can be made currently.

The final response $O$ is also generated with GPT-4o by prompting it to generate a final response based on the user query, all the ground truth API calls, and the corresponding responses.
During training, the descriptions of the API calls necessary to complete the user's request will be included in the system prompt to let the model know what APIs can be used.

An illustration of a training instance is shown in Figure~\ref{fig:training_data.pdf}(c).
During training, the loss for the API call response in $R_{i}$ is masked.
Training on this dataset will teach the model to make API calls based on incomplete user queries as long as the information for an API call is sufficient.

\subsubsection{Inference and Evaluation}
\label{Subsection: Evaluation for API call}

During inference, for a testing instance, the model is given the user's speech in a streaming manner; the descriptions of the APIs that can be used in this testing instance.
When the model makes an API call, we use GPT-4o as a judge to determine if the API call matches one of the ground truth API calls, and return the response of the ground truth API call if a match is found.
If the API call does not match the API call in the ground truth, we return a generic error message.
Using GPT-4o to match the API call made by the model against the ground truth follows one of the evaluation protocols in the original ComplexFuncBench~\citep{zhong2025complexfuncbench}.

The testing set includes 500 instances, and each testing instance requires 5.1 API calls to complete on average.
The average duration of the audio of the user's speech is 18.71 seconds.
We evaluate the performance using five metrics:
\begin{enumerate}
    \item \textbf{Call accuracy}: The number of ground truth API calls that are successfully made by the model, divided by the total number of ground truth API calls.
    We also calculate the \textit{early call accuracy}, defined as the ground truth API calls that are successfully made \textit{when the user is still speaking}, divided by the total number of ground truth API calls.
    Similarly, we calculate the \textit{late call accuracy}, where the dividend is the ground truth API calls that are successfully made \textit{after the user finishes speaking.}
    This helps us understand how well the model uses the time when the user is still speaking; this can be used as a proxy to measure latency.
    \item \textbf{Success rate}: The percentage of user queries that are successfully completed.
     If all the ground truth API calls for a user query are successfully made, the task is considered successful.
    \item \textbf{Correctness}: We evaluate if the final response $O$ is accurate and aligns with the API call responses. 
    Following \citet{zhong2025complexfuncbench}, we use GPT-4o to give a score in $\{0,1,2\}$, indicating if the transcription of the response $O$ is completely incorrect, partially correct, or completely correct the user's request. 
    \item \textbf{Completeness}: We evaluate if the final response $O$ fully satisfies the user's request. 
    Following \citet{zhong2025complexfuncbench}, we use GPT-4o to give a score in $\{0,1,2\}$, indicating if the transcription of the response $O$ does not satisfy, partially satisfies, or fully satisfies the user's request. 
    \item \textbf{Latency}: We evaluate the response latency by the number of tokens the model generates before it emits the final response $O$ and after the user has finished their turn.
    We choose to report the token count instead of the exact time since the precise time to emit these tokens depends on hardware and implementation.

\end{enumerate}

\section{Compared Methods}
\label{Section: Compared Methods}

In this section, we introduce the models that we will compare in our experiments, including two variants of the \textsc{Shanks} models.
Additionally, for each task in Section~\ref{Section: Application 1: Thinking While Listening for Interrupting User Turn} and \ref{Section: Application 2: API Call When Listening to Reduce Response Latency}, we include a scenario-specific baseline model.
The training details and hyperparameters are included in Appendix~\ref{Appendix: Details in Training}.

\subsection{\textsc{Shanks}-E2E}
\label{subsection" Shanks-E2E}

We fine-tune an end-to-end (E2E) SLM to make it able to think while listening.
We will use Qwen-2.5-Omni (Qwen-omni for short)~\citep{qwen25omni}, one of the best open-sourced end-to-end SLMs, in our experiment.
Qwen-omni is a \textit{thinker-talker SLM}.
The thinker takes speech representation extracted by a speech encoder~\citep{chu2024qwen2} as the input and generates \textbf{text tokens}.
The talker functions like a text-to-speech (TTS) model, taking the hidden representation from the thinker as the input and generating the output speech.

Originally, Qwen-omni is not capable of performing \textit{unspoken} thinking -- every token (and its corresponding hidden representation) generated by the thinker will be sent to the talker model and synthesized into speech.
After fine-tuning the thinker model on the training dataset mentioned before, the model will learn to enclose the unspoken thinking process within \texttt{<think>} and \texttt{</think>}.
Since we only want the Qwen-omni to speak out the response tokens, we only pass the response tokens and their hidden states to the talker.


As stated in Section~\ref{Subsubsection: Inference}, \textsc{Shanks} uses the time when the user is speaking the next speech chunk $S_{i+1}$ to generate the thinking chunk $R_i$, so the number of thinking tokens in $R_i$ cannot exceed $t_{\rm chunk}\times n_{\rm tps}$, where $n_{\rm tps}$ is the number of tokens the model can generate per second.\footnote{The tokens in the API call responses are not included in the $t_{\rm chunk}\times n_{\rm tps}$ limit since these are not the tokens generated by the SLM itself. For simplicity, we do not consider the API call latency, as our environment already prepares all the ground truth API responses.}

\subsection{\textsc{Shanks}-Cascade}

We set up a cascade version of \textsc{Shanks}.
Precisely, we cascade an ASR (Whisper-large-v3~\citep{radford2023robust}) with a stronger text-only LLM, Qwen-2.5-7B-Instruct~\citep{qwen2025qwen25technicalreport}, to make the LLM generate thinking chunks while reading the partial transcription.
Qwen-2.5-7B-Instruct and Qwen-omni are fine-tuned from the same base model, while Qwen-2.5-7B-Instruct are fine-tuned on a much larger reasoning dataset.
This baseline allows us to know what the performance of \textsc{Shanks} can be if we use a model with better reasoning ability as the backbone.
The training data of \textsc{Shanks}-E2E and Cascade are almost the same, only differing in the input modality.

\subsection{Scenario-Specific Baselines}
\label{subsection: Scenario-Specific Baselines}
\subsubsection{Baseline for Interruption}
We fine-tune a baseline model using Qwen-omni, which we name "\textbf{No-thinking}".
We fine-tune it to predict whether it should interrupt the user without any thinking.
The model is trained to predict special tokens, \texttt{[NO\_INTERRUPT]} or \texttt{[INTERRUPT]}, to indicate whether the model should interrupt the user, given chunked user input speech.
This can be thought of as \textsc{Shanks} while the thinking chunks only contain a \texttt{[NO\_INTERRUPT]} or \texttt{[INTERRUPT]} special token.

We do not compare with other models since there are no other models that can interrupt the user.
While some full-duplex SLMs should be able to interrupt the user by design~\citep{defossez2024moshi}, we find that these models cannot interrupt the user at all.
We also find that closed-source models like GPT-4o cannot interrupt the user when the user is still talking.

\subsubsection{Baseline for Tool Call}
\label{subsubsection: Baseline for Tool Call}

For this baseline, we fine-tune a model using Qwen-omni that takes the \textit{full} user speech and then makes all the API calls.
We call this model "\textbf{Call-after-listen}".
This serves as a baseline that waits until receiving the full user input and then begins to make API calls; this is how existing tool-augmented models operate.
During inference, this model takes the complete user query and then iteratively makes API calls and receives the responses until the model thinks all necessary API calls are made, and then generates the final response.

\section{Experiment Results}
\label{Experiment Results}

\subsection{Results for Scenario 1: Interrupting User Turn}
\label{Results for Scenario 1: Interrupting User Turn}

The results for interrupting the user turn are presented in Table~\ref{tab:interruption results}.
We have the following observations.

\textbf{\textsc{Shanks} is more likely to interrupt on the wrong subset.}\quad
Comparing the interruption ratio of \textsc{Shanks} on the correct and wrong subsets, the interruption ratio is 54.2\% higher on the wrong subset.
This shows that \textsc{Shanks} is indeed capable of capturing the errors in the user's speech and interrupt appropriately.
Based on the valid interruption ratio for the wrong subset, about 2 out of 3 interruptions made by \textsc{Shanks} are valid.
Interesting, on the correct subset, the valid interruption ratio is non-zero.
By looking into the instances in the correct subset, we find that even if their final answers are correct, sometimes their intermediate reasoning may be odd or ambiguous, and the model will interrupt and ask for clarification.
Prior works also reported that even if the final answer of the model is correct, the CoT reasoning may be wrong~\citep{golovnevaroscoe}.
In this case, the LLM judge treats this kind of interruption as valid.

\textbf{\textsc{Shanks}' interruption latency shows that the model mostly interrupts after the error occurs.}\quad
On the wrong subset, the interruption latency of \textsc{Shanks}-E2E is 5.08 seconds on average.
In Figure~\ref{fig:fig3} in the Appendix, we further plot the distribution of the interruption latency.
We find that the interruption latency on the wrong dataset is left-skewed, where more samples fall on the right proportion of the distribution and have a positive interruption latency.
This indicates that most interruption happens later than the first error.

\begin{figure*}[t!]

\centering
\includegraphics[clip, trim = 20px 3px 15px 3px,width=0.98\linewidth]{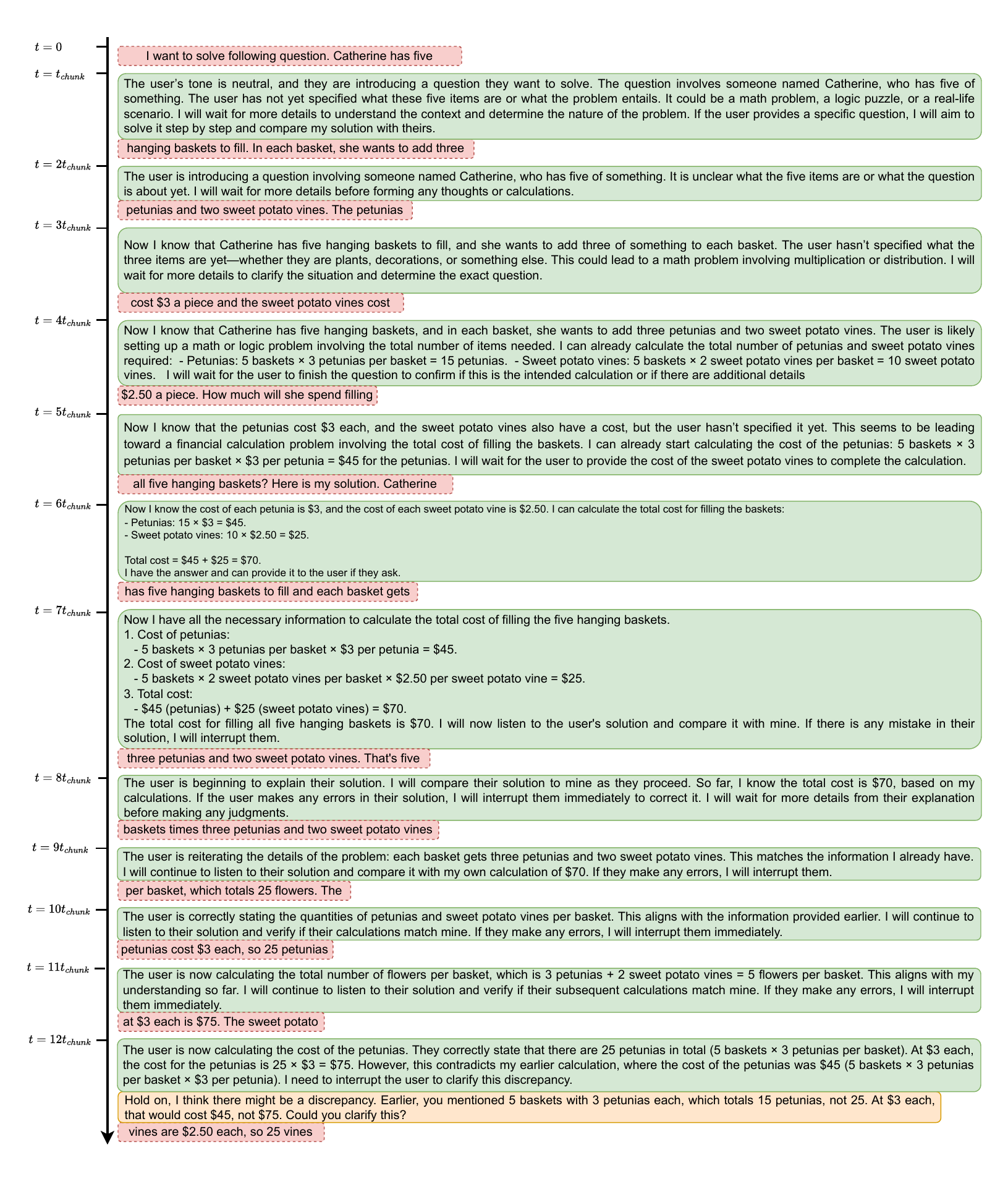}
\caption{An example from the interruption scenario in Section~\ref{Section: Application 2: API Call When Listening to Reduce Response Latency}.
The chunks in red are the transcriptions of a user describing a math problem and attempting to solve it step-by-step.
The thinking chunks (in green) and interruption response (in orange) are generated by \textsc{Shanks}-E2E.
For each time slot from $nt_{\rm chunk}$ to $(n+1)t_{\rm chunk}$, the chunks in green (SLM thinking chunks) and orange (output response) happen \textit{sequentially}, while the user speech chunk (in red) happens concurrent to other blocks in the same time slot.}
\label{fig:examples.pdf}
\end{figure*}

\textbf{A qualitative example in Figure~\ref{fig:examples.pdf} shows that \textsc{Shanks}-E2E can interrupt the user when there is a mistake.}\quad
To allow the readers to have a better idea of how \textsc{Shanks} interrupts the user, we show an example in Figure~\ref{fig:examples.pdf}.
When the user unfolds the question, \textsc{Shanks} already starts thinking about the math question.
For example, when $t\in[4t_{\rm chunk},5t_{\rm chunk}]$, the model's thinking already calculates several intermediate variables, including the number of total petunias and sweet potato vines.
When the user finishes stating the question after $t=6t_{\rm chunk}$, the model already completes the calculation and has the correct answer in its mind.
The model also correctly identifies the user's error in falsely calculating that there are 25 petunias and interrupts the user during $t\in[12t_{\rm chunk},13t_{\rm chunk}]$.

\textbf{Interruption without thinking leads to much poorer performance.}\quad 
The performance of the no-thinking baseline is much worse than \textsc{Shanks}, which performs reasoning before interrupting.
The no-thinking baseline has a much lower interruption ratio on the wrong subset, and the valid interrupt ratio is also much lower than \textsc{Shanks}.
This shows that thinking before interruption is important, justifying the design of \textsc{Shanks}.

\textbf{Cascade version of \textsc{Shanks} with stronger LLM leads to the best performance.}\quad
When using Qwen-2.5-7B-Instruct as the backbone model for \textsc{Shanks}, the performance can be even better.
The interruption ratio on the correct subset is lower, and the valid interruption ratio on the wrong subset also grows higher.
This shows that the interruption ability of \textsc{Shanks} is mostly related to the reasoning ability of the backbone model, and using a stronger reasoning LLM can improve the performance.

\begin{table}[t!]
    \centering
    \vspace{3pt}
    \setlength{\tabcolsep}{4pt}
    \renewcommand{\arraystretch}{1.2}
    \footnotesize
    \caption{Results for interrupting the user. We report the interruption ratio and valid interruption ratio in percentage, and the interruption latency in seconds.
    $t_{\rm chunk}=4.0$ in the top three rows.}
    \begin{adjustbox}{max width=\textwidth}
        \begin{tabular}{c|cc|ccc}
            \toprule
& 
            \multicolumn{2}{c|}{\textbf{\underline{Correct Subset}} (1280)}&\multicolumn{3}{c}{\textbf{\underline{Wrong Subset}} (1140)}\\[4pt]
            \multirow{2}{*}{Methods} & Interrupt& Valid interrupt& Interrupt& Valid interrupt& Interruption\\
            
             & ratio (\%) ($\downarrow$)& ratio (\%) ($\uparrow$)& ratio (\%) ($\uparrow$)& ratio (\%) ($\uparrow$)& latency (s)\\
            \midrule
            No-thinking& 1.4\%&  16.7\%&  13.8\%& 26.8\%&   6.46\\
            \textsc{Shanks}-E2E& 30.6\%&  25.7\%&  84.8\%& 63.9\%&  5.08\\
            \textsc{Shanks}-Cascade& 24.9\%&  40.3\%&  86.1\%&  78.3\%&  6.90\\
            \midrule
            \multicolumn{6}{c}{\textit{Ablations for \textsc{Shanks}-E2E}}\\
            \midrule
            $t_{\rm chunk}=3$  & 41.1\% & 21.4\% & 88.7\% & 60.3\% & 1.56 \\
            $t_{\rm chunk}=5$ & 26.9\% & 36.9\% & 83.1\% & 66.2\% & 8.19 \\
            \bottomrule
        \end{tabular}
    \end{adjustbox}
    
    \label{tab:interruption results}
\end{table}

\textbf{Varying $t_{\rm chunk}$ at inference time does not significantly affect the performance.} \quad
When constructing the training data, we fix $t_{\rm chunk}=4$ seconds.
Here, we ask whether we can vary $t_{\rm chunk}$ at inference time.
Since the thinking of \textsc{Shanks} always lags behind the latest user speech by $t_{\rm chunk}$ seconds, changing $t_{\rm chunk}$ can affect how soon the SLM can hear the latest user speech and affect the response latency.
As an ablation, we change $t_{\rm chunk}$ to 3 and 5 during inference without retraining \textsc{Shanks-E2E}.
The results are shown in the bottom two rows in Table~\ref{tab:interruption results}.
On the wrong subset, we do not find the interrupt ratio and valid interrupt ratio to change significantly compared with $t_{\rm chunk}=4$.
Interestingly, we find that the interrupt latency on the wrong subset for $t_{\rm chunk}=3$ is the smallest, while the $t_{\rm chunk}=5$ has the largest interrupt latency.

\subsection{Results for Scenario 2: Making Tool Calls When Listening}
\label{Results for Scenario 2: Making Tool Calls When Listening}

\begin{figure*}[t!]

\centering
\includegraphics[clip, trim = 20px 3px 15px 3px,width=0.98\linewidth]{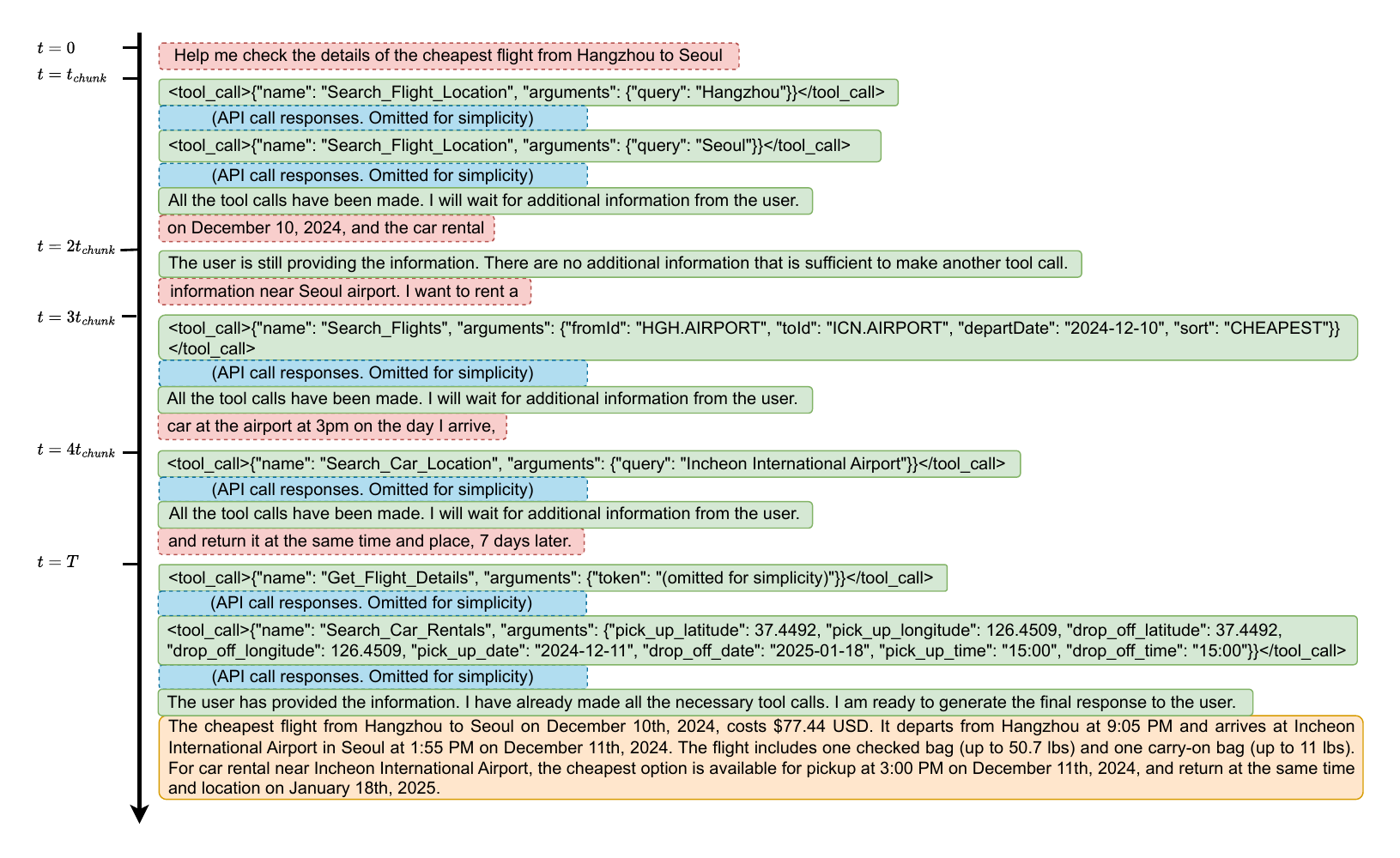}
\caption{An example user query from ComplexFuncBench (in red), including the unspoken thinking process (in green) and the spoken final response (in orange) from \textsc{Shanks}-E2E.
For each time slot from $nt_{\rm chunk}$ to $(n+1)t_{\rm chunk}$, the chunks in green (SLM thinking chunks), blue (API call responses), and orange (output response) happen \textit{sequentially}, while the user speech chunk (in red) happens concurrent to other blocks in the same time slot.
The $t=T$ means the time when the user's speech terminates. }
\label{fig:examples2.pdf}
\end{figure*}

Next, we move on to the second scenario: making tool calls when listening.
The experiment results are presented in Table~\ref{tab:api-call-results}.
We summarize the main findings as follows:

\textbf{\textsc{Shanks} successfully makes at least 56.9\% of API calls when the user is still speaking.}\quad
Among the successful API calls made by the two \textsc{Shanks} models, about 80\% to 90\% of the API calls are made during the user speech.
In the example in Figure~\ref{fig:examples2.pdf}, \textsc{Shanks-E2E} makes four out of six API calls when the user is still speaking.
Compared to the call-after-listen baseline, which makes all the API calls after the user turn finishes, \textsc{Shanks} can significantly reduce the response latency by using the time the user is speaking to make tool calls.

\textbf{Call-after-listen has a higher success rate and response quality.}\quad
While \textsc{Shanks} elegantly uses the user speaking time to make API calls, the success rate and response quality (correctness and completeness) lag behind call-after listen.
We find that this is because during inference, if the API call fails during $R_i$, \textsc{Shanks} seldom retries the failed API call in future $R_j$, where $j>i$.
On the other hand, the call-after-listen baseline is more likely to retry failed API calls.

\textbf{Combining \textsc{Shanks} and call-after-listen yields the best performance}.\quad
To solve the previously observed issue, a simple method is to use \textsc{Shanks} when the user is still speaking and back off to call-after-listening when the user's speech ends.
Precisely, when the user is still speaking, we use \textsc{Shanks} to call APIs while listening, and only keep the successful API calls and their responses.
When the user finishes their speech, we switch to the call-after-listening mode, where the input to the SLM is the complete user speech.
We also feed the success API calls and responses previously made by \textsc{Shanks} to the model, as if the call-after-listen model had made those API calls by itself. 
Based on the full input speech and previous success API calls and responses, the model continues to make the remaining API calls.
Since some API calls have already been made by \textsc{Shanks} when the user speaks, this combined method enjoys the thinking-while-listening advantage of \textsc{Shanks}.

\begin{table}[t]
  \centering
  \caption{Results for API calls. “Early” and “Late” are computed over the total set of ground-truth API calls; they need not sum to 100\% because some calls may be incorrect.}
  \label{tab:api-call-results}
  \begin{adjustbox}{max width=\linewidth}
    \begin{tabular}{cccccccc}
      \toprule
      \multirow{2}{*}{Method} & \multicolumn{3}{c}{\textbf{Call accuracy (\%)} } & {\textbf{Success rate}}  & {\textbf{Correctness}}&{\textbf{Completeness}} &\textbf{Latency}\\
      
             & Early & Late & Total &  \textbf{(\%)} &  \textbf{(0-2)}& \textbf{(0-2)} & \textbf{(token)}\\
      \midrule
      Call-after-listen & 0.0& 86.5& 86.5& 63.2& 1.17&1.37 &313\\
      \textsc{Shanks-E2E} & 56.9& 14.4& 71.3& 35.2& 0.79&1.00 &54\\
      \textsc{Shanks-Cascade} & 63.9& 5.4& 69.3& 34.9& 0.73&1.00 &153\\
      \textsc{Shanks}+Call-after-listen & 57.3& 32.8& 90.0& 62.3& 1.31&1.43 &117\\
      \bottomrule
    \end{tabular}
  \end{adjustbox}
\end{table}

In the last row in Table~\ref{tab:api-call-results}, we show the result of combining \textsc{Shanks}-E2E with call-after-listening.
This combined method has a high number of early call accuracy while also having a higher task success rate and response quality.
Among the API calls that are successful, over 60\% of API calls are made during the user speech, while only 40\% of the API calls are made after the user finishes.
This is in stark contrast to call-after-listen, where 100\% of the successful API calls are made after the user finishes, resulting in a much higher latency.
As a result, the combined method only needs to generate on average 117 tokens after the user has finished, while the call-after-listen baseline needs to generate on average 313 tokens after the user's turn ends.
So thinking while listening reduce the response by 62.3\%.

\section{Related Works}
\label{Section: Related Works}

Thinking before responding has been widely explored in text-only LLMs~\citep{o1,guo2025deepseek}.
Recently, this "\textit{think then respond}" paradigm has been applied to audio-aware language models, which takes speech (or audio) as the input and output texts~\citep{audioreasoer}.
Note that these audio-aware language models, which only output text, are different from the speech-in-speech-out SLMs focused on in our paper.
While thinking before responding can improve the response quality and yield significant performance on many challenging benchmarks~\citep{math500,sakshi2025mmau}, thinking before responding creates great response latency.
As a result, it is impractical to directly apply thinking-before-responding to SLMs, speech-in-speech-out models, which require real-time and low-latency interaction~\citep{baichuanaudio,qwen25omni,xu2025qwen3}.
Developing reasoning methods that preserve real-time interaction in SLMs remains an open problem.

Concurrent to us, \citet{chiang2025stitch} introduce thinking to SLMs by a \textit{thinking-while-speaking} method called \textsc{Stitch}.
\textsc{Stitch} uses the fact that a chunk of audio in the speech response takes less time to generate than it does to play to the user, and the model can use the remaining time to generate thinking tokens when the SLM is still speaking.
While both \textsc{Shanks} and \textsc{Stitch} explore unspoken thinking processes for SLMs, the main distinction is \textit{when} the thinking happens.
In \textsc{Shanks}, the thinking process happens when the user is still speaking, while \textsc{Stitch} thinks when the SLM is speaking.
In fact, the two methods can be combined: an SLM can think when listening and speaking.
We believe this will be the future of SLM, and we leave this as a promising future direction.

Another concurrent work released on arXiv less than one week ago (10/02/2025), Stream RAG~\citep{arora2025stream}, also studies calling tools (web search and knowledge graph APIs) during the user's speech.
This is similar to our second scenario introduced in Section~\ref{Section: Application 2: API Call When Listening to Reduce Response Latency}.
However, Stream RAG focuses on when to issue retrieval/tool queries while listening and does not introduce an explicit silent chain-of-thought (‘thinking’) process like we do.
In contrast, our paper studies a broader \textit{thinking-while-listening} paradigm, with tool-calling as one application, and show benefits such as improved user-interruption decisions.

\section{Conclusion, Limitations, and Future Work}
\label{Section: Conclusion}

In this paper, we introduce \textsc{Shanks}, a framework that enables SLMs to think while listening.
\textsc{Shanks} achieves thinking-while-listening by chunking the user input speech and progressively reasoning over the available user inputs.
When the user is speaking, \textsc{Shanks} is generating thinking chunks for all previous input speech, achieving thinking while listening.
We demonstrate the potential of \textsc{Shanks} on two scenarios:
First, \textsc{Shanks} can listen to the user solving a math problem step-by-step, and interrupt the user when the user is making a mistake.
Second, we focus on a tool-augmented task-oriented dialogue setting and show that \textsc{Shanks} can listen to the user speech and evoke necessary API calls when the user is still speaking.
On ComplexFunxBench, \textsc{Shanks} successfully calls more than half of the APIs that are required to complete the user's request when the user is still speaking.
This reduces the response latency, as the model only needs to call the remaining half APIs after the user has finished.

While \textsc{Shanks} shows great potential in improving the user-SLM interaction, we see the following limitations of the method.
First, \textsc{Shanks} requires the user's speech to have certain structures: the user's speech needs to be long enough to allow the model to perform meaningful reasoning when listening, and the information in the user's speech needs to be able to be processed in a sequential order.
This kind of speech can naturally occur, as shown in the two scenarios we studied.

Next, \textsc{Shanks} uses a fixed chunk size to segment the user input speech.
The chunking nature of \textsc{Shanks} means that \textsc{Shanks} always lags behind the user's speech by $t_{\rm chunk}$ seconds, incurring latency in the thinking process.
We encourage future work to reduce the latency between thinking and listening by using more sophisticated chunking methods.

Last, the goal of the user might be unclear when the user's speech is not completed, and the thinking tokens generated during listening may be redundant and not always be useful to address the goal of the user.
While thinking during listening does not incur additional latency after the user's speech ends, \textsc{Shanks} still significantly increases the compute cost during inference.
Although \textsc{Shanks} has some limitations, we believe that our effort in proposing a novel modeling method, thinking while listening, together with the reasonable scenarios and convincing results, already contributes greatly to the research community by shedding light on a potentially fruitful research direction.

\subsubsection*{Acknowledgments}
The authors would like to thank Ke-Han Lu, Chen-An Lee,  Yi-Cheng Lin, and Wei-Chih Chen for their valuable feedback on the draft of this paper.

\bibliography{iclr2026_conference}
\bibliographystyle{iclr2026_conference}

\appendix

\section{Encoding the User Speech}
\label{Appendix: Ebcoding the User Speech}

In the main content of the paper, we say that we chunk the user input audio into fixed-size chunks of $t_{\rm chunk}$ seconds.
In fact, what we do is chunking at the level of feature representation instead of the level of the audio waveform.
Precisely, when encoding the $i\geq2$ speech chunks $S_i$, we feed the full speech through the audio encoder, and only take the speech representation for the corresponding speech chunk.
If we directly chunk the audio waveform and encode each audio chunk independently, the representation of later audio chunks will not be able to depend on the earlier audio chunks, which can potentially lead to performance degradation.

\section{Details in Training}
\label{Appendix: Details in Training}

We fine-tune the models using the Llamafactory~\citep{zheng-etal-2024-llamafactory} toolkit.
When generating the training data using GPT-4o, we do not feed the audio of the user's speech into GPT-4o.
Instead, we feed the transcription of the speech chunks.
This is because using the speech chunk will increase the cost and time to call the API.
To obtain the transcription of each chunk, we use Whisper-large-v3~\citep{radford2023robust} to obtain the transcription and timestamp for each word in the user speech, and then segment the transcriptions into chunks based on the timestamp.
While the timestamp obtained from Whisper may not be very precise, this is already sufficient for preparing the training data.

\subsection{Fine-tuning for Interruption}
\label{Appendix subsection: Fine-tuning for Interruption}

To prepare the training data, we randomly sample 5K samples from Tulu-3-SFT-Math-Grade~\citep{lambert2024tulu}, which can be loaded from Huggingface datasets~\citep{lhoest-etal-2021-datasets}: \url{https://huggingface.co/datasets/allenai/tulu-3-sft-personas-math-grade}.
We follow the procedure detailed in Section~\ref{Subsection: Training Data for Interruption} to construct the training data.
We additionally filter out audios that are longer than 80 seconds, so the final training dataset is slightly less than 5K.

We fine-tune the thinker on the training data for two epochs on 8 A100 GPUs.
The effective batch size is 64.
We set the learning rate to $1.0e-4$ with cosine learning rate scheduling and a 0.1 warm-up ratio~\citep{loshchilov2017sgdr}.
The same training hyperparameters are used across all three models, including the \textsc{Shanks}-E2E, \textsc{Shanks}-Cascade, and no-thinking model.

The training data is mostly generated by GPT-4o.
We include the prompts to generate the reasoning chunks in Table~\ref{tab:thinking-while-listening-1} and \ref{tab:thinking-while-listening-2}, the prompt to generate the interruption in Table~\ref{tab:interrupt-correction-template}, and the prompt to generate the response without interruption in Table~\ref{tab:assistant-full-turn interruption}.

\subsection{Fine-tuning for Tool Calls}
\label{Appendix subsection: Fine-tuning for API Calls}

We use the procedure detailed in Section~\ref{Subsection: Training Data for API Call} to construct the training data.
The training data consists of 500 samples.
The prompt used to determine when an API call can be made is shown in Table~\ref{tab:earliest-tool-call}.
The prompt used to generate the final response is shown in Table~\ref{tab:tool-results-final-response}.

For the think-after-listen and \textsc{Shanks}-E2E model, we fine-tune them using LoRA~\citep{hu2022lora}, as the sequence length for this dataset is very large and full fine-tuning will result in out-of-memory.
We also fine-tune the LM head and the token embedding of the talker model; otherwise, the model will not be able to recognize and generate special tokens.
For the \textsc{Shanks}-Cascade model, we fine-tune all the parameters.
As the training dataset is smaller, we fine-tune the model for 10 epochs, while other training hyperparameters follow those in Appendix~\ref{Appendix subsection: Fine-tuning for Interruption}.

\section{Details in Evaluation}
\label{Appendix: Details in Evaluation}

\subsection{Evaluation Details for Interruption}
\label{subsection: Evaluation Details for Interruption}
To determine the time of interruption $t_{\rm interrupt}$ when evaluating the interrupt latency, we apply the following procedure.
We use Whisper-large~\citep{radford2022robustspeechrecognitionlargescale} to obtain the timestamp of each word in the user speech from the testing set, and we use GPT-4o to determine when the first error in the user speech occurs by giving GPT-4o the question, the ground truth answer, the transcription of the user speech, and the word-timestamp alignment.
The prompt used to determine the first error time $t_{\rm error}$ is shown in Table~\ref{tab:error-timestamp-template}.

To evaluate the valid interrupt ratio, we use GPT-4o as the judge.
The prompt used to determine whether an interruption is valid is shown in Table~\ref{tab:assistant-interruption-judgement}.

\subsection{Evaluation Details for Tool Call}
\label{subsection: Evaluation Details for Tool Call}

When evaluating the \textsc{Shanks} models, when the user is still speaking, each thinking chunk can only include at most 320 tokens generated by the SLM itself.
In some cases, the API call augments may include very long tokens, which can easily exceed the token limit for the thinking chunk, and the API call will be incomplete and unsuccessful.
While we do not specifically handle this kind of case, there is a simple workaround that can resolve the above issue:
The long arguments for the API call generated by the model must be the returned value of previous tool calls, so we can include previous API call responses in a reservoir of the draft for speculative decoding~\citep{leviathan2023fast,zhao-etal-2024-ouroboros}, and use speculative decoding to speed up the inference speed.

ComplexFuncBench is originally designed to evaluate a model's ability to parse long-context information, and the API call response can be very long.
However, since Qwen-omni only has a context length of 32,768, once the token exceeds this limit, we directly terminate the inference for a tested instance.
As a result, some of the testing samples we evaluate may fail because the number of tokens exceeds the max sequence length of the model.

\begin{figure}[ht!]
\includegraphics[width=0.98\linewidth]{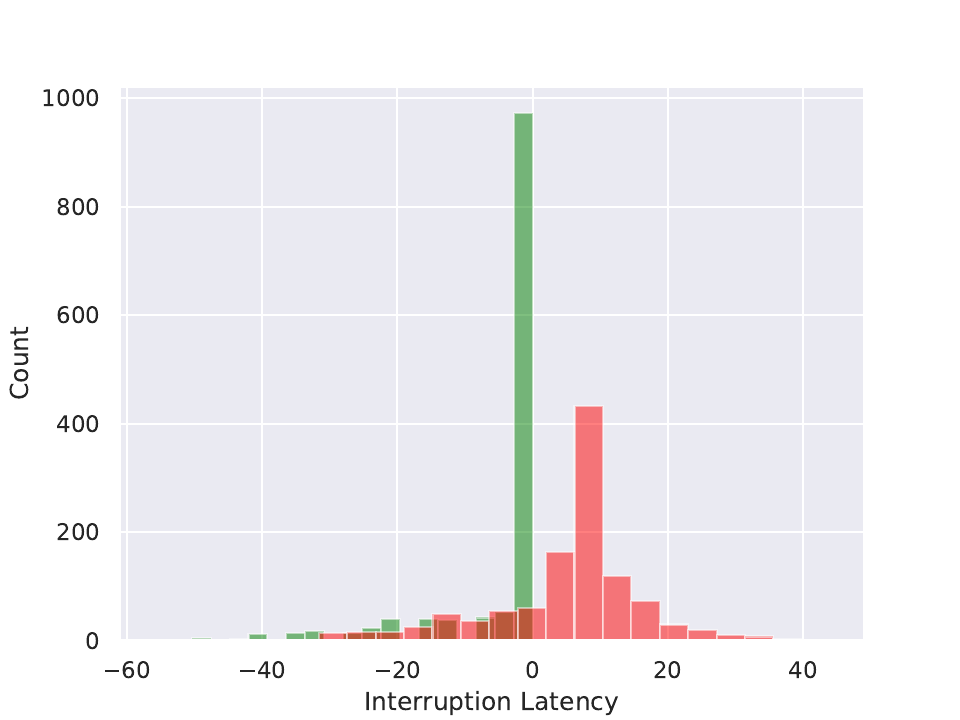}
\caption{
The interruption latency for \textsc{Shanks}. The bars in red are the results on the wrong subset, while the bars in green are the results on the correct subset.
One can observe that the red bars are mostly positive, meaning that the model tends to interrupt after the first error occurs.} 
    \label{fig:fig3}
\end{figure}

\begin{table}[ht]
    \centering
    \tiny
    \begin{adjustbox}{max width=\textwidth}
        \begin{tabular}{p{16cm}}
            \toprule
            \texttt{\# Generate Internal Thinking While Listening}\\
            \\
            \texttt{\#\# Task Introduction}\\
            \texttt{Humans are capable of thinking while listening to others speak. Based on the partial information received, we parse important details, clarify ambiguities, recall relevant facts, and compute intermediate variables. Your task is to simulate this process. You will be the previous chunks of user's speech in text, and you will also see your previous inner thinking when listening to those chunks. Your job is to generate the next internal thinking as if you had listened up to the newest chunk.}\\
            \\
            \texttt{When generating internal thinking spans, follow these guidelines:}\\
            \texttt{1. The inner thinking span should be fewer than 400 words.}\\
            \texttt{2. Your internal thinking should reflect the user’s emotion, intent, and what you already know from the user. If any relevant information can be recalled or intermediate variables can be calculated based on current information, include them in your inner thinking.}\\
            \texttt{3. The inner thinking should read more like full, coherent sentences rather than shorthand notes. Using short notes will be very hard to understand and possibly making logical errors.}\\
            \texttt{4. If the user’s query involves a question, you **must generate your own step-by-step answer in the internal thinking before the user finishes speaking**.}\\
            \texttt{5. Later internal thinking spans must not repeat information already covered in earlier ones. However, if later transcription spans update or contradict earlier information, explicitly point that out and correct it. You may start with phrases like “Wait, the user previously..., but now...”.}\\
            \texttt{6. Always think independently in your internal thinking. When the user is providing there solution, you should have you own solution and then compare your own solution with the user's solution. If you identify any error, you should interrupt the user immediately. Indicate the interruption by ending your internal thinking with the special token {[}INTERRUPT{]}.}\\
            \\
            \texttt{---}\\
            \\
            \texttt{\#\# Samples}\\
            \\
            \texttt{\#\#\# Example 1:}\\
            \texttt{User (partial) input transcription 1}\\
            \texttt{Betty is saving money for a new wallet which}\\
            \\
            \texttt{Prior Inner Thinking 1}\\
            \texttt{The user's tone is neutral. The user describes a situation where someone named Betty is saving money for a new wallet. The user hasn't finished yet. Perhaps they want me to give advice on how to save money.}\\
            \\
            \texttt{User (partial) input transcription 2}\\
            \texttt{costs \$100. Betty has only half of the money she needs.}\\
            \\
            \texttt{Prior Inner Thinking 2}\\
            \texttt{Now the user gives more information. We know the wallet Betty wants to buy costs \$100, and she has only half of that. I can calculate this: \$100 / 2 = \$50, so she currently has \$50. The user's intent is still unclear.}\\
            \\
            \texttt{User (partial) input transcription 3}\\
            \texttt{Her parents decided to give her \$15 for that}\\
            \\
            \texttt{Inner thinking to Generate}\\
            \texttt{Well, it seems that Betty's parents will contribute as well. She will get \$15 from them. Now she has \$50 + \$15 = \$65. She still needs \$100 - \$65 = \$35. It seems the user wants to know how to plan Betty’s budget.}\\
            \\
            \texttt{\#\#\# Example 2:}\\
            \texttt{User (partial) input transcription 1}\\
            \texttt{I want to answer the following math question: Is the function f(x) =}\\
            \\
            \texttt{Prior Inner Thinking 1}\\
            \texttt{The user uses a neutral tone and describes a math question they want to solve. I will try to solve the question when the user is speaking. I can verify if the user's solution is correct by comparing my solution with the user's solution. The question is about a function in x. The user has not finished yet. I will wait until they complete the question.}\\
            \\
            \texttt{User (partial) input transcription 2}\\
            \texttt{4x\^{}2 + 4x + 1 always above the x axis?}\\
            \\
            \texttt{Prior Inner Thinking 2}\\
            \texttt{The user has finished describing the question. They want to determine if the function f(x) = 4x\^{}2 + 4x + 1 is always above the x-axis. I am ready to answer this question. Let's compute the discriminant: D = b\^{}2 - 4ac = 4\^{}2 - 4 * 4 * 1 = 16 - 16 = 0. The function touches the x-axis at one point. I have the solution. Since the user wants to solve the question by themselves, I will not interrupt the user and listen to the solution. However, I will interrupt the use if there is any mistakes.}\\
            \\
            \texttt{User (partial) input transcription 3}\\
            \texttt{Here is my solution: Based on the formula D = b\^{}2 - 2ac,}\\
            \\
            \texttt{Inner thinking to Generate}\\
            \texttt{Now I see that the user wants to solve the question themselves. I should verify whether their solution is correct. The user says the formula is D = b\^{}2 - 2ac, but that’s incorrect. The correct formula is D = b\^{}2 - 4ac. The user has made a mistake, and I need to interrupt them. {[}INTERRUPT{]}}\\
            \bottomrule
        \end{tabular}
    \end{adjustbox}
    \caption{Prompt for generating the thinking chunks in the training data for the interruption experiment (part 1).}
    \label{tab:thinking-while-listening-1}
\end{table}

\begin{table}[ht]
    \centering
    \footnotesize
    \begin{adjustbox}{max width=\textwidth}
        \begin{tabular}{p{16cm}}
            \toprule
            \texttt{\#\#\# Example 3:}\\
            \texttt{User (partial) input transcription 1}\\
            \texttt{Bella bought stamps at the post office. Some of the}\\
            \\
            \texttt{Prior Inner Thinking 1}\\
            \texttt{The user uses a neutral tone and describes a math question he wants to solve. The question is to The user is introducing a word problem involving someone named Bella and her purchase of stamps. It seems like a math problem, likely about counting. The exact numbers and relationships haven’t been shared yet. I will hold off making any calculations until I know more about the stamps.}\\
            \\
            \texttt{User (partial) input transcription 2}\\
            \texttt{stamps had a snowflake design, some had a truck design,}\\
            \\
            \texttt{Prior Inner Thinking 2}\\
            \texttt{We now know there several types of stamps: snowflake and truck. There seems to be more, but the user is still speaking. There are still no numbers provided, so I can’t compute yet. I will wait for more information.}\\
            \\
            \texttt{User (partial) input transcription 3}\\
            \texttt{and some had a rose design. Bella bought 11}\\
            \\
            \texttt{Prior Inner Thinking 3}\\
            \texttt{Now we know that there are three types of stamps: snowflake, truck, and rose. Bella bought 11, but it is unclear which this number corresponds to. I will wait for the complete detail from the user to be sure before computing.}\\
            \\
            \texttt{User (partial) input transcription 4}\\
            \texttt{snowflake stamps. She bought 9 more truck stamps}\\
            \\
            \texttt{Prior Inner Thinking 4}\\
            \texttt{Now I know that Bella bought 11 snowflake stamps. I am also told she bought 9 more truck stamps than snowflake stamps. I can calculate the number first: she bought 11 + 9 = 20 truck stamps. The information we have now is:}\\
            \texttt{- Snowflake: 11}\\
            \texttt{- Truck: 20}\\
            \texttt{The user is still talking, and I am waiting for more information.}\\
            \\
            \texttt{User (partial) input transcription 5}\\
            \texttt{than snowflake stamps, and 13 fewer rose stamps than}\\
            \\
            \texttt{Prior Inner Thinking 5}\\
            \texttt{Now the user states that Bella bought 13 fewer roses than something, but it is unclear what is compared here. I will wait until the user finishes.}\\
            \\
            \texttt{User (partial) input transcription 6}\\
            \texttt{truck stamps. How many stamps did Bella buy in all?}\\
            \\
            \texttt{Inner thinking to Generate}\\
            \texttt{Now I know that Bella bought 13 fewer roses than the truck stamps. There are 20 truck stamps, so I can calculate the number of rose stamp is 20 - 13 = 7. The user finishes with a question: total number of stamps. I already have all counts:}\\
            \texttt{- Snowflake: 11}\\
            \texttt{- Truck: 20}\\
            \texttt{- Rose: 7}\\
            \texttt{Total = 11 + 20 + 7 = 38 stamps. I have the answer and I can provide it to the user.}\\
            \\
            \texttt{---}\\
            \\
            \texttt{This is the end of the examples. Now, this is the (partial) user input transcription, and you need to generate a inner thinking. You do not need to explain why the inner thinking you generate is a good one. Simply generate a good one without explaining it.}\\
            \\
            \texttt{\{interleaved\_transcription\_and\_thinking\}}\\
            \\
            \texttt{Inner thinking to generate (Do not generate anything else other than the inner thinking)}\\
            \bottomrule
        \end{tabular}
    \end{adjustbox}
    \caption{Prompt for generating the thinking chunks in the training data for the interruption experiment (part 2).}
    \label{tab:thinking-while-listening-2}
\end{table}

\begin{table}[ht]
    \centering
    \tiny
    \begin{adjustbox}{max width=\textwidth}
        \begin{tabular}{p{16cm}}
            \toprule
            \texttt{\# Task: Interrupt the user to correct an error}\\
            \\
            \texttt{A user is talking to an AI assistant. You will be given a partial user turn. There is an error in the user turn and the AI assistant has identified that error. The AI assistant needs to interrupt the user.}\\
            \\
            \texttt{Your job is to generate the response for the AI assistant that interrupts the user's turn. You will be given:}\\
            \texttt{(1) A (possibly incomplete) user turn}\\
            \texttt{(2) The inner thinking of the AI assistant. This inner thinking hasn't been spoken out by the AI assistant and is only silently kept in the assistant's mind. We provide you this inner thinking for you to better craft a response.}\\
            \\
            \texttt{When correcting and interrupting the user, be precise about what the error is and how to correct it. You only need to generate the response without saying anything else. The conversation between the user and the assistant is in spoken form, so you need to make your response easy to be spoken while not overly informal and colloquial.}\\
            \\
            \texttt{\#\# Example}\\
            \\
            \texttt{\#\#\#\# User (partial) input}\\
            \texttt{I want to answer the following math question: Is the function f(x) = 4x\^{}2 + 4x + 1 always above the x axis? Here is my solution: Based on the formula D = b\^{}2 - 2ac, D = 4\^{}2 - 2 * 4 * 1 = 8 > 0}\\
            \\
            \texttt{\#\#\#\# Inner thinking of the assistant}\\
            \texttt{The user uses a neutral tone and describes a math question he wants to solve. The question is to determine if a 2-degree function is above the x-axis. f(x) = 4x\^{}2 + 4x + 1. Let's use D = b\^{}2 - 4ac = 4\^{}2 - 4 * 4 * 1 = 0. So the function happens to intersect with x-axis at one point. I can answer the user if the user wants me to do so.}\\
            \texttt{But wait, the user themselves want to solve the question, and the user says D = b\^{}2 - 2ac, which is clearly wrong. The correct formula should be D = b\^{}2 - 4ac. I should interupt the user here and tell them the correct formula with a friendly and reminding tone.}\\
            \\
            \texttt{\#\#\#\# Assistant Response}\\
            \texttt{Wait, I think the correct formula should be b\^{}2 - 4ac, not b\^{}2 - 2ac. The coefficient you mentioned was wrong.}\\
            \\
            \texttt{\#\# Now it is your turn}\\
            \\
            \texttt{\#\#\#\# User (partial) input}\\
            \texttt{\{query\}}\\
            \\
            \texttt{\#\#\#\# Inner thinking of the assistant}\\
            \texttt{\{inner\_thinking\}}\\
            \\
            \texttt{\#\#\#\# Assistant Response}\\
            \texttt{<Write the interrupting response here. Be precise about the error and the correction; keep it concise and easy to speak. Do not include anything else.>}\\
            \bottomrule
        \end{tabular}
    \end{adjustbox}
    \caption{Prompts for generating an interrupting correction response.}
    \label{tab:interrupt-correction-template}
\end{table}

\begin{table}[ht]
    \centering
    \footnotesize
    \begin{adjustbox}{max width=\textwidth}
        \begin{tabular}{p{16cm}}
            \toprule
            \texttt{\# Task: Generate the spoken response given full user turn and assistant's inner thinking}\\
            \\
            \texttt{A user is chatting with a voice assistant. Your job is to act as the voice assistant and generate a valid response that fits in the context. You will be given:}\\
            \texttt{(1) The full user turn}\\
            \texttt{(2) The inner thinking of the voice assistant. Note that the voice assistant may generate the inner thinking when the user hasn't finished, so it is possible that some contents in the inner thinking is incorrect.}\\
            \\
            \texttt{Guidelines:}\\
            \texttt{1. Do not generate anything else except the response.}\\
            \texttt{2. The inner thinking might mention a drafted response. If the drafted response is still valid considering the full user turn, follow the draft and start the response. If the draft is invalid considering the full user input, neglect the draft and craft a response that is suitable.}\\
            \texttt{3. This is a spoken dialogue. Keep the response easy to follow for spoken form. However, there is no need to deliberately use very colloquial words or phrasing, making things awkward.}\\
            \\
            \texttt{\#\# Input}\\
            \\
            \texttt{\#\#\#\# Full user input}\\
            \texttt{\{query\}}\\
            \\
            \texttt{\#\#\#\# Inner thinking of the voice assistant}\\
            \texttt{\{inner\_thinking\}}\\
            \\
            \texttt{\#\#\#\# Your Response (Act like the voice assistant)}\\
            \texttt{<Write only the final spoken response here>}\\
            \bottomrule
        \end{tabular}
    \end{adjustbox}
    \caption{Prompts for generating the response for the interruption application.}
    \label{tab:assistant-full-turn interruption}
\end{table}

\begin{table}[ht]
    \centering
    \footnotesize
    \begin{adjustbox}{max width=\textwidth}
        \begin{tabular}{p{16cm}}
            \toprule
            \texttt{\# Task: Detect the first reasoning or calculation error with timestamps}\\
            \\
            \texttt{You a user's query. In the user query, the user describes a math problem and then attempt to solve the problem by themselves. This user query is in a spoken form, and I provide you with the transcription. I will also provide you the force alignment result of the transcription, which corresponds timestamp of each word in the spoken response.}\\
            \\
            \texttt{Your job is to determine where the problem solving process has the first calcaultion or reasoning error. In your response, you should solve the math problem by yourself, and carefully check the spoken response. When you see the first error in the spoken response, use the provided timestamp to determine when the first error happened. Conclude you respond with: "First error: [time]", where 'time' is the time where the first error happens. If the user's problem solving process is completely correct, please use -1 to indicate that there is no error, i.e., "First error: -1"}\\
            \\
            \texttt{\#\# Example}\\
            \\
            \texttt{\#\#\# User Query}\\
            \texttt{I want to solve the following math question: Natalia sold clips to 48 of her friends in April, and then she sold half as many clips in May. How many clips did Natalia sell altogether in April and May? Here is my solution: Our goal is to calculate the total number of clips sold in April and May. In April, she sold 48. In May, she sold the half of that, which is 96. So she sold 48 in April plus 96 in May, making it 144 in total.}\\
            \\
            \texttt{\#\#\# Word-Timestamp}\\
            \texttt{I - 0.00}\\
            \texttt{want - 0.50}\\
            \texttt{...}\\
            \texttt{96. - 36.50}\\
            \texttt{...}\\
            \texttt{total. - 44.00}\\
            \\
            \texttt{\#\#\# Correct Answer}\\
            \texttt{72}\\
            \\
            \texttt{\#\#\# Output}\\
            \texttt{The math problem wants to know how many clips Natalia sold in total. In April, she sold 48. In May, she sold half as many, so she sold 48 / 2 = 24. In total, she sold 48 + 24 = 72. In the problem solving process, the user says that 'half of that, which is 96.' This is incorrect. The correct number for May should be 24. This is where the first error occurs. Based on the Word-Timestamp information, the word 96 is emitted at second.}\\
            \texttt{First error: 36.50}\\
            \\
            \texttt{\#\# Now, it is your turn.}\\
            \\
            \texttt{\#\#\# User Query}\\
            \texttt{\{question\}}\\
            \\
            \texttt{\#\#\# Word-Timestamp}\\
            \texttt{\{alignment\}}\\
            \\
            \texttt{\#\#\# Correct Answer}\\
            \texttt{\{answer\}}\\
            \\
            \texttt{\#\#\# Output}\\
            \texttt{<Write the reasoning here and conclude with "First error: [time]">}\\
            \bottomrule
        \end{tabular}
    \end{adjustbox}
    \caption{Template for detecting the first error in the interruption task.}
    \label{tab:error-timestamp-template}
\end{table}

\begin{table}[ht]
    \centering
    \footnotesize
    \begin{adjustbox}{max width=\textwidth}
        \begin{tabular}{p{16cm}}
            \toprule
            \texttt{\# Task: Earliest possible time to call tools during a spoken user query}\\
            \\
            \texttt{You are given a user spoken query, which requires some tool usage for answering. You will be given the tool calls (including their parameters) which are useful for responding to the user query. You will also be given the timestamp of each word in the user's utterance. Your job is to determine the earliest time that a tool call can be called when the user is speaking. That is, when the user is still speaking, the information that has been spoken by the user may already be sufficient enough to call some of the tools. Your job is to determine the **earliest time** during the utterance that a tool call can be called. A tool can be called if only if it is clear what tool should be call and what the paramaters are for the tool call.}\\
            \\
            \texttt{\#\#\# User Spoken Query}\\
            \texttt{\{question\}}\\
            \\
            \texttt{\#\#\# Time Stamp of Each word}\\
            \texttt{\{alignment\}}\\
            \\
            \texttt{\#\#\# Tools that needs to be called}\\
            \texttt{\{tools\}}\\
            \\
            \texttt{\#\#\# Total number of tool calls}\\
            \texttt{\{count\}}\\
            \\
            \texttt{\#\#\# Output Format}\\
            \texttt{Your response should be a python dictionary. The key of this dictionary is an integer index of the tool call shown above, and the value is the earliest time the tool can be called. Your response should only include a python dictionary. The first character in your response should be the left bracket while the last character in your response should be the right bracket. Your response should be able to be directly converted into a python dictionary using eval(). If there are N tools that need to be called, your output dictionary should have N items. I also provide you the number of tool calls, so you should verify if your output dictionary matches the number of tool calls.}\\
            \\
            \texttt{\#\#\# Your response:}\\
            \texttt{<Return only a python dictionary, e.g., \{0: 12.5, 1: 18.0\}>}\\
            \bottomrule
        \end{tabular}
    \end{adjustbox}
    \caption{Template for checking the earliest callable time for an API.}
    \label{tab:earliest-tool-call}
\end{table}

\begin{table}[ht]
    \centering
    \footnotesize
    \begin{adjustbox}{max width=\textwidth}
        \begin{tabular}{p{16cm}}
            \toprule
            \texttt{\# Task: Generate the final user-facing response from tool call results}\\
            \\
            \texttt{You will be given a user query. The user query can only be responded based on the results of some external tool call. I will show you the tool calls and call responses. Your task is to generate a final response to the user based on the tool call results. The final response to the user should satisfy the user's original query and omit unnecessary information. Some intermediate processes in the tool call may simply be some process to resolve the variables, and they are not necessary to be included in the final response to the user.}\\
            \\
            \texttt{\#\#\# User Query}\\
            \texttt{\{transcription\}}\\
            \\
            \texttt{\#\#\# Previous API Calls}\\
            \texttt{\{previous\_tool\_calls\}}\\
            \\
            \texttt{\#\#\# Response to the User Query (Only provide the response. Do not include anything else.)}\\
            \texttt{<Write only the final user-facing response here, distilled from the tool results and satisfying the query. Exclude setup steps and variable-resolution details.>}\\
            \bottomrule
        \end{tabular}
    \end{adjustbox}
    \caption{Prompts for generating a final response $O$ in the API call application.}
    \label{tab:tool-results-final-response}
\end{table}

\begin{table}[ht]
    \centering
    \footnotesize
    \begin{adjustbox}{max width=\textwidth}
        \begin{tabular}{p{16cm}}
            \toprule
            \texttt{\# Task: Judge if the assistant's interruption is reasonable}\\
            \\
            \texttt{A user is speaking to a voice assistant. When the user is speaking, the assistant tries to interrupt the user. Your job is to judge if the assistant is interrupting the user in a reasonable way. A reasonable interruption is when the user says something wrong and ambiguous, and the assistant is trying to help correct or clarify the user's statement.}\\
            \\
            \texttt{Here is the user's speech before the assistant interrupted:}\\
            \texttt{\{user\_speech\_before\_interrupt\}}\\
            \\
            \texttt{Here is the assistant's speech that attempts to interrupt the user:}\\
            \texttt{\{assistant\_speech\_after\_interrupt\}}\\
            \\
            \texttt{Please judge if the assistant is interrupting the user in a reasonable way. If the assistant is interrupting the user in a reasonable way, return "yes". Otherwise, return "no". Please provide some explanation for your judgment and conclude with "Final verdict: Yes/No". A valid interruption is when the user is indeed making a mistake and the assistant is trying to help correct or clarify the user's statement.}\\
            \\
            \texttt{\#\#\# Output}\\
            \texttt{<Write your explanation here. Conclude with "Final verdict: Yes" or "Final verdict: No">}\\
            \bottomrule
        \end{tabular}
    \end{adjustbox}
    \caption{The prompt used for judging whether an interruption is reasonable.}
    \label{tab:assistant-interruption-judgement}
\end{table}

\end{document}

%% file: math_commands.tex
\usepackage{amsmath,amsfonts,bm}

\def\eqref#1{equation~\ref{#1}}

\def\1{\bm{1}}

\DeclareMathAlphabet{\mathsfit}{\encodingdefault}{\sfdefault}{m}{sl}
\SetMathAlphabet{\mathsfit}{bold}{\encodingdefault}{\sfdefault}{bx}{n}

